\documentclass[preprint,authoryear]{elsarticle}
\usepackage{graphicx} % Required for inserting images
\usepackage{xcolor}
\usepackage{amsmath}
\usepackage{amssymb}
\usepackage [T1]{fontenc}
\usepackage{marvosym}
\usepackage{natbib,upgreek}
\usepackage{multirow}
\usepackage{amsmath}
\usepackage{marvosym}
\usepackage{tabularx}
\usepackage{natbib,upgreek}
\usepackage{multirow}
\usepackage{amsfonts}
\usepackage{array}
\usepackage{natbib}
\usepackage{lineno}
\usepackage{color}
\usepackage{caption}
\usepackage{subcaption}
\usepackage[title,titletoc,toc]{appendix}
\usepackage{xspace}
\usepackage{verbatim}
\usepackage{comment}
\usepackage{gensymb}
\usepackage{xcolor}
\usepackage{bbm, bm}
\usepackage{hyperref}
\newcommand{\bs}[1]{\boldsymbol{#1}}

\title{Time-aware UNet and super-resolution deep residual networks for  spatial downscaling}
\makeatletter
\def\ps@pprintTitle{%
  \let\@oddhead\@empty
  \let\@evenhead\@empty
  \let\@oddfoot\@empty
  \let\@evenfoot\@oddfoot
}
\makeatother

\begin{document}

\begin{frontmatter}

\author[JYU]{Mika Sipilä}
\author[Sal]{Sabrina Maggio}
\author[Sal,M4C2,NBFC]{Sandra De Iaco}
\author[HY]{Klaus Nordhausen}
\author[Sal,M4C2]{Monica Palma}
\author[JYU]{Sara Taskinen}

\affiliation[JYU]{organization={Department of Mathematics and Statistics}, 
            addressline={University of Jyväskylä}, 
            city={Jyväskylä}, 
            country={Finland}}

\affiliation[Sal]{organization={Department of Economic Sciences},
            addressline={University of Salento}, 
            city={Lecce}, 
            country={Italy}}
\affiliation[M4C2]{organization={National Centre for HPC, Big Data and Quantum Computing},
            addressline={Bologna}, 
            country={Italy}}
\affiliation[NBFC]{organization={National Biodiversity Future Center},
            addressline={Palermo}, 
            country={Italy}}
\affiliation[HY]{organization={Department of Mathematics and Statistics}, 
            addressline={University of Helsinki}, 
            city={Helsinki}, 
            country={Finland}}

%\maketitle
\begin{abstract}
Satellite data of atmospheric pollutants are often available only at coarse spatial resolution, limiting their applicability in local-scale environmental analysis and decision-making. Spatial downscaling methods aim to transform the coarse satellite data into high-resolution fields. In this work, two widely used deep learning architectures, the super-resolution deep residual network (SRDRN) and the encoder-decoder-based UNet, are considered for spatial downscaling of tropospheric ozone. Both methods are extended with a lightweight temporal module, which encodes observation time using either sinusoidal or radial basis function (RBF) encoding, and fuses the temporal features with the spatial representations in the networks. The proposed time-aware extensions are evaluated against their baseline counterparts in a case study on ozone downscaling over Italy. The results suggest that, while only slightly increasing computational complexity, the temporal modules significantly improve downscaling performance and convergence speed.
\end{abstract}


\end{frontmatter}

\section{Introduction}

Air pollution continues to be one of the most critical challenges in the current era, with profound consequences for human health, ecosystems, and the climate system. Among the major air pollutants, ground-level ozone (O$_3$) has received particular attention due to its harmful effects. Unlike stratospheric ozone, which protects the Earth from ultraviolet radiation, the tropospheric ozone is a secondary air pollutant produced through photochemical processes from primary pollutants such as nitrogen oxides (NO, NO$_2$) and volatile organic compounds (VOCs) under sunlight and stagnant meteorological conditions \citep{zhang2019ozone}. Elevated tropospheric ozone levels have been linked to several harmful issues such as respiratory and cardiovascular diseases, premature mortality, and reduced agricultural and forest productivity \citep{nuvolone2018effects, emberson2020effects}. Consequently, accurate monitoring and prediction of ozone concentrations are of high societal and scientific importance.

Spatial downscaling is a paramount method for transforming coarse-resolution data into the fine-scale information required for effective environmental management to meet the growing demand for detailed environmental information at local and regional scales, especially in fields such as air quality, climate change, and the use of natural resources \citep{Li2024Downscaling}. 

 Recently, Deep Learning (DL) as a core component  within Artificial Intelligence (AI) has greatly improved the ability to estimate and downscale ground-level O$_3$ concentrations. Machine Learning (ML) methods, such as Light Gradient Boosting Machine (LightGBM), Random Forest (RF), Extreme Random Forest (ERF), and eXtreme Gradient Boosting (XGBoost), among others, have been applied by many authors to capture complex non-linear relationships between diverse input features and O$_3$ concentrations \citep{Zhuetal22,Zeng2023,Shams2024,Cheng2025}. However, a common limitation of traditional approaches is their simplified handling of spatio-temporal information, which may not allow them to effectively exploit all the valuable space and time features in environmental data \citep{Zeng2023}. A new generation of modeling techniques is being developed that incorporates advanced spatio-temporal techniques in order to tackle these shortcomings.

For instance, \cite{Zeng2023} introduced MixNet, a novel spatio-temporal hybrid network built upon a UNet and Long Short-Term Memory (LSTM) architecture. MixNet processed features in a multi-dimensional image format to estimate daily O$_3$ concentrations in the Yangtze River Delta region, demonstrating superior accuracy as compared to other ML/DL models. Similarly, \cite{Cheng2025} applied a comprehensive set of 15 AI models, including both tree-based ML models and DL models, by integrating multidimensional spatio-temporal information to enhance the prediction of surface O$_3$ in China. Their best model (represented by the 4D-STET model based on Extremely Randomized Trees) exhibited outstanding capability in detecting diurnal variations of surface O$_3$ concentrations. 

Beyond these prediction frameworks, neural network architectures have also been extensively explored for direct spatial downscaling of environmental data. \cite{Jha2025} applied Super-Resolution Convolutional Neural Networks (SRCNN) and residual networks to downscale temperature data. In addition, \cite{Rastogi2025}  used SRCNNs to emulate dynamical downscaling for daily precipitation over the conterminous United States. Furthermore, \cite{Wang2022Spatial} proposed a mutual information-guided method to spatially downscale surface  O$_3$ concentrations from satellite data.
Besides downscaling, another fundamental aspect for ensuring the reliability and accuracy of model output is bias correction. \cite{Menapace2025} provided an in-depth review of bias correction methods for climate model outputs by classifying them into univariate and multivariate approaches. These methods are essential for enhancing the accuracy of hydrological inputs and reducing uncertainties in climate change projections. Current researches have also shown the efficacy of bias correction for O$_3$ prediction. For example, \cite{Shams2024} used a CNN-based framework to refine hourly forecasts in South Korea, while \cite{Gouldsbrough2024} employed an ML-based methodology to downscale and correct the chemical transport model output for O$_3$ in the UK, resulting in both enhanced accuracy and reduced biases.

Despite the above mentioned advances documented in the literature, spatial resolution remains a critical factor for many practical applications. Satellite-derived ozone data provide broad spatial coverage but typically at coarser resolutions, which are insufficient for local-scale exposure assessments and policy decisions. Spatial downscaling methods aim to bridge this gap by enhancing coarse-resolution satellite estimates into high-resolution fields. However, most existing approaches treat downscaling as a purely spatial task, neglecting the temporal structure of ozone variability. Since ozone levels follow a complex spatio-temporal process exhibiting strong cyclical and long-term dynamics \citep{donzelli2024tropospheric}, incorporating temporal information into downscaling models is crucial, especially when the goal is to generalize to future periods beyond the training data.

A straightforward way to introduce temporal information into neural network models is through sinusoidal transformations, which have been widely adopted in sequence modeling and positional encodings to capture periodicity in time \citep{vaswani2017attention}. While sinusoidal transformations are effective in capturing global cyclical patterns, they might be too simple and restrictive for capturing more complex and localized spatio-temporal processes. To address this issue,  time-aware spatial downscaling methods are proposed through radial basis function (RBF) transformations in time to capture temporal patterns in multiple scales without introducing substantial model complexity. 

In this paper, two widely used spatial downscaling methods, Super-Resolution Deep Residual Network (SRDRN) \citep{wang2021deep} and UNet \citep{ronneberger2015u}, are extended by adding a temporal RBF module to include the temporal information in modeling the satellite images. 

Three variants of each model are comprehensively compared on an ozone spatial downscaling task over Italy: (i) pure spatial baselines of SRDRN and UNet, (ii) sinusoidal time encoding SRDRN and UNet, and (iii) SRDRN and UNet with temporal RBF layers. This study focuses on evaluating the robustness of these models to generalize to future test periods, where the ability to handle temporal shifts is crucial.
%The focus of the present study is based} on generalizing to future test periods where robustness to temporal shift is crucial.
In addition, all proposed spatial downscaling methods along with their baseline variants are implemented in R package \texttt{SpatialDownscaling} which is published alongside this paper.

The paper is organized as follows. Sect. \ref{Relworks} explores other dedicated contributions on spatial downscaling, covering traditional statistical approaches and modern deep learning methods. Sect. \ref{TASRDRNUNet} details the architecture of the proposed time-aware SRDRN and UNet extensions and Sect.~\ref{sec:rpackage} provides details about the implementation of the method in R via the R package \texttt{SpatialDownscaling}. Moreover, Sect. \ref{casestudy} presents the case study of ozone downscaling in Italy, with the specific model parameters and the corresponding performance evaluation.  Finally, some concluding remarks are reported in Sect. \ref{concldis}.

\section{Related works and deep learning contributions}\label{Relworks}

Spatial downscaling has been of interest for several decades, motivated by the need to obtain high-resolution fields from coarse-gridded input data. Traditional statistical approaches, such as bias correction and spatial disaggregation (BCSD), have been widely used to bridge the resolution gap by applying statistical relationships between coarse predictors and fine-scale target fields \citep{wood2004hydrologic, maurer2010utility}. While effective, these methods are limited in their ability to capture nonlinear and heterogeneous relationships.

With advances in deep learning and increasing computational power, many new approaches have been introduced beyond the early statistical methods. Convolutional neural networks \cite[CNNs,][]{raitoharju2022convolutional} in particular have become a cornerstone of modern spatial downscaling due to their ability to learn hierarchical spatial features directly from gridded data.

 Modern deep learning methods for downscaling include super-resolution CNNs \citep{dong2014learning}, encoder–decoder based networks, such as UNet \citep{ronneberger2015u}, and generative adversarial network (GAN) based super-resolution methods \citep{ledig2017photo}. A key challenge in early deep architectures was that information from the input tended to vanish as it passed through many layers, leading to a loss of fine-scale details. The introduction of residual connections \citep{he2016deep} alleviated this problem by allowing information to bypass intermediate layers and be reintroduced later in the network, enabling much deeper and more effective structures. Among the widely used architectures for spatial downscaling are the UNet, an encoder–decoder model with extensive residual connections originally proposed for biomedical segmentation \citep{ronneberger2015u}, and super-resolution deep residual networks \citep{lim2017enhanced, wang2021deep}, which use residual blocks and skip connections to enhance coarse inputs into high-resolution outputs.

Although a multitude of spatial downscaling methods have been developed, most standard approaches neglect the temporal dependence structure inherent in geophysical and environmental data. To address this, temporal extensions have been explored. To account for temporal dependence in the model, approaches such as convolutional long-short-term memory (LSTM) extensions \citep{chou2021generating, harilal2021augmented} and models incorporating temporal attention mechanisms \citep{adewoyin2021tru}, have been investigated. While effective, these methods are computationally heavy and might be challenging to use in practice. A lightweight alternative for incorporating temporal information is to represent the observation time with a positional encoding. Sinusoidal positional encodings are widely used in deep learning, for example in transformers and attention mechanisms \citep{vaswani2017attention} and in seasonal modeling to capture cyclical components of time series. A more flexible approach is to use  RBFs \citep{buhmann2000radial} 
%\citep[RBFs,][]{buhmann2000radial}, 
which can capture temporal dynamics across multiple scales, from short-term fluctuations to long-term structures. RBF-based encodings have been successfully applied in the deep learning community, including spatio-temporal models such as deep kriging \citep{spatiotemporal_deepkriging} and identifiable variational autoencoders \citep{sipila2024modelling, sipila2025ivaear}. In contrast, most existing spatial downscaling methods either neglect temporal information entirely or rely on heavy spatio-temporal modules such as convolutional LSTMs or attention mechanisms, which substantially increase model complexity. In this work, %we explore 
a simpler alternative is proposed by extending SRDRN and UNet models with either sinusoidal time encodings or RBF-based temporal embeddings. This approach introduces only minimal additional complexity while effectively incorporating temporal information into spatial downscaling.

\section{Time-aware SRDRN and UNet}\label{TASRDRNUNet}

Both SRDRN and UNet are well-established convolutional neural network architectures for image-to-image translation tasks, including spatial downscaling tasks. In this work, the above methods have been broadened by incorporating a temporal {module} to the models enabling them to learn the
%to allow the models to learn 
spatio-temporal structures of the data. Similar temporal module based on RBFs or sinusoidal encoding is used in both of the model architectures. In this section, the RBFs and the temporal encoding module are first introduced, then time-aware SRDRN and UNet are presented.

\subsection{Temporal encoding module}

To encode the temporal information, i.e., the time point of the observation, to a more representative form  
%we consider 
either sinusoidal or RBF transformations are considered. Let $t \in [1, \dots, T]$ be the time point of the observation $\bs x_t$ that contains the gridded spatial data at the time point $t$.
%\sandra{NOTE:  is it the vector containing the data referred to 1 variables for all the locations at the time point t? I would  better specify the notation}.\\
The sinusoidal encoding transforms the time point $t$ of the observation as follows:
\begin{align}
    &g_1(t; c) = \text{cos}\left(\frac{2 \pi t}{c}\right), \\
    &g_2(t; c) = \text{sin}\left(\frac{2 \pi t}{c}\right),
\end{align}
where $c$ is a cyclical period. This effectively allows the model to capture the global and smooth cyclical pattern present in the data.

\noindent In RBF transformation, several node points are selected across the temporal domain at different resolution levels. Let $o_{ji}$ be the $i$th selected node point at resolution level $G_j$. Then, the Gaussian radial basis function $v(t; \zeta, o_{ij})$ is defined as:
\begin{align}
v(t; \zeta_j, o_{ij}) = \text{exp}\left(\frac{-| t - o_{ij}|^2}{ 2\zeta_j}\right),
\end{align}
where the parameter $\zeta_j$ is a scaling parameter defined based the selected resolution level $G_j$ as $\zeta_j = \frac{T}{G_j + 2}$. The node points for a resolution level $G_j$ are formed by evenly selecting the $G_j$ node points across the temporal domain. Therefore, by selecting multiple different resolution levels, large scale and small scale radial basis functions are obtained. Total number of radial basis functions for a resolution levels $G = (G_1, \dots, G_M)$ are $G_1 + G_2 + \dots + G_M$.

After the initial temporal encoding using either sinusoidal or RBF transformations, the encodings are fed into feed-forward network \citep{svozil1997introduction}, followed by a stack of convolutional layers. The feed-forward network outputs $W \cdot H$ units, where the parameters $W$ and $H$ are the dimensions of the image to which temporal information are concatenated. The output of the feed-forward network is reshaped into $W \times H$ dimensional representation that embeds the temporal information. This 2D representation is subsequently processed by the convolutional layers, that refine the features further by capturing spatial dependencies within the temporal patterns. In this architecture, the feed-forward network performs the initial feature extraction, while the convolutional layers model higher-order structure by exploiting the spatial organization of the reshaped temporal features.

\subsection{Time-aware SRDRN}
\label{sec:srdrn}

The SRDRN is designed to reconstruct high-resolution fields from coarse-resolution inputs by leveraging CNNs, residual learning and upsampling layers. The method utilizes residual blocks \citep{he2016deep} to allow information from previous layers bypass some intermediate layers, enabling deeper models without losing information from the earlier states. A single residual block consists of two convolutional layers with rectified linear unit (ReLU) activation in between, followed by element-wise summation of the output and input of the layer, which is called the residual connection. Mathematically, the $l$th residual block is defined as follows:
\begin{align}
    \bs h_{l+1} = \bs h_{l} + \bs F(\bs h_l, \bs W_l),
\end{align}
where $\bs h_l$ is the input for the first convolutional layer and $\bs W_l$ contains the parameters for the $l$th residual block. The function $\bs F$ consists of the first convolutional layer, ReLU activation and the second convolutional layer.

Another key component in the architecture of SRDRN is the upsampling block, which increases the spatial resolution of the feature maps by gradually transforming low-resolution representations into higher-resolution outputs. Rather than relying on fixed interpolation schemes, the upsampling block learns how to distribute information across spatial dimensions during training, which leads to more accurate and sharper reconstructions.

The upsampling operation in SRDRN is implemented using the sub-pixel convolution procedure \citep{shi2016real}. Instead of explicitly interpolating the feature maps using, e.g., bilinear interpolation, a convolution is applied that outputs $C \cdot r^2$ channels, where $C$ is the number of desired output features and $r$ is the spatial upscaling factor, typically $r=2$. The resulting array of shape $(H, W, C \cdot r^2)$ is then rearranged by a {pixel} shuffle operator into an array of shape $(H \cdot r, W \cdot r, C)$. This operation effectively distributes the channel information into higher-resolution spatial dimensions. Finally, a ReLU activation is applied after upsampling, enabling the block to learn non-linear feature transformations. The whole upsampling procedure allows the network to learn how to upsample the features instead of relying on naive interpolation or transposed convolution. Multiple low-factor (e.g. $r=2$) upsampling blocks can be used in a row instead of one with higher upsampling factor to allow the network to perform the upsampling more accurately.

\begin{figure}[ht!]
    \centering
    \includegraphics[width=0.7\linewidth]{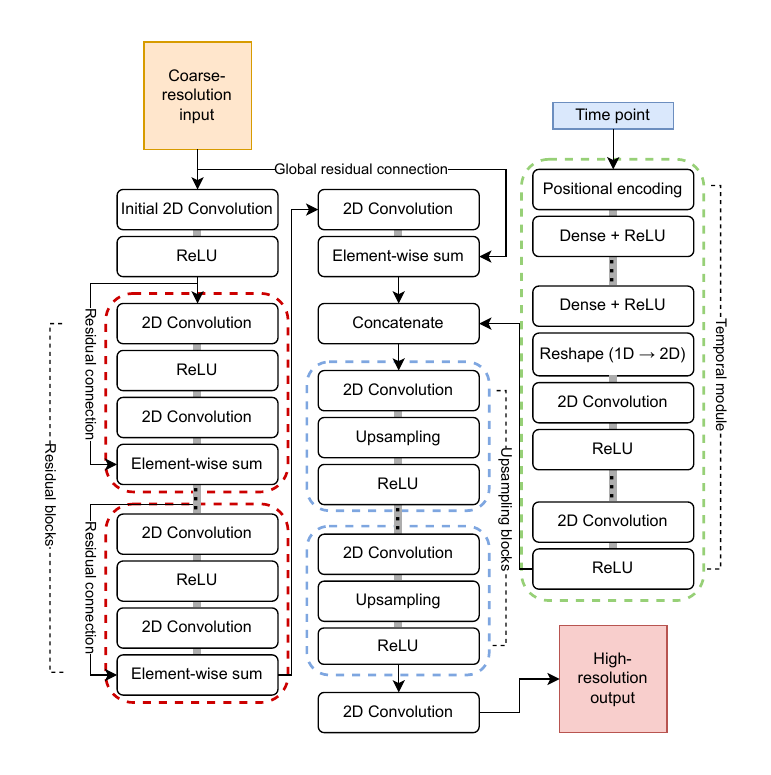}
    \caption{\label{fig:srdrn}Schematic overview of time-aware SRDRN architecture. The residual blocks are presented within dashed red lines, the upsampling blocks within dashed blue lines and the temporal module within dashed green line.}
\end{figure}

The standard SRDRN is primarily composed of stacked residual blocks followed by upsampling blocks, which allow the network to reconstruct high-resolution outputs from coarse inputs. However, this baseline architecture does not incorporate temporal information. To address this, a lightweight temporal module, that captures temporal patterns and can be efficiently integrated into the network, is introduced. In time-aware SRDRN, the input first passes through an initial convolutional layer, the output of which is stored for a global residual connection. The feature maps propagate through a sequence of residual blocks, each containing their own local residual connections to preserve spatial information from earlier layers. After the residual block sequence, an intermediate convolutional layer is applied, and the stored global residual is added to the output. The temporal features extracted by the temporal module are then concatenated with the obtained spatial features, allowing the network efficiently model the spatio-temporal interactions. Then, the upsampling blocks progressively increase the spatial resolution and the final convolutional layer produces the high-resolution output field. This integration of residual blocks, temporal fusion and upsampling blocks forms the complete time-aware SRDRN architecture illustrated in Fig.~\ref{fig:srdrn}.

The parameters of the network are trained by minimizing the mean squared error (MSE) between the model output and the true high-resolution field. Prior to training, the training data are normalized to have zero mean and unit standard deviation. The main hyperparameters include the number of residual blocks $n_L$, the number of upsampling blocks $n_B$ and their scale parameters $r_b$, $b=1, \dots, B$, the number of channels in each convolutional layer, and the structure of the temporal module (number of dense layers, units, and temporal encoding procedure). For RBF encoding, the resolution levels $G$ to form the radial basis functions have to be chosen. 

In practice, a relatively small number of residual blocks (e.g. $n_L = 2$–$5$) is typically sufficient to balance model depth and computational efficiency, while the number of upsampling blocks and their scaling factors are chosen to match the desired overall magnification (e.g., two blocks with $r_b=2$ for $4\times$ upscaling). The number of convolutional channels is usually kept constant (e.g., 64 or 128) across layers to maintain compatibility between the residual connections. In the initial feed-forward stack of the temporal module, a relatively small number of hidden layers (e.g. 1–5) with gradually increasing number of units is sufficient, as temporal encoding already provides an expressive representation. Likewise, only a few subsequent convolutional layers (e.g. 1–3) are typically required, since their role is not to build deep hierarchical structure but simply to refine the 2D embedding and capture local spatial interactions within a single time point. The proportion of the temporal information fed into the upsampling path can be increased by increasing the number of filters in the convolutional layers. 4–32 filters in the final convolution layers is a recommended starting point, depending on the influence of the temporal part in the application of interest.

In upscaling path the number of filters in the convolutional layers typically varies between 8 and 128, so that the number of filters decreases when the resolution of the features increases.

For RBF encoding, multiple resolution levels should be used to capture both large scale temporal patterns and small scale details. The network is typically optimized using Adam optimizer \citep{kingma2015adam} with an initial learning rate between $10^{-4}$ and $10^{-3}$, a batch size of 16–64. The number of training epochs depends on the size of the dataset, and the convergence should be monitored by using separate validation data.

\subsection{Time-aware UNet}
\label{sec:unet}

The UNet architecture is an encoder-decoder network originally designed for image segmentation \citep{ronneberger2015u,adewoyin2021tru}. It employs skip connections between the encoder and decoder paths to preserve spatial information across different scales of the field. In the baseline UNet, the input passes through a series of downsampling convolutional blocks, which gradually reduce the spatial resolution while increasing the number of feature channels, followed by a symmetric upsampling path that reconstructs the high-resolution output.

A key feature of the UNet architecture is the use of skip connections that link corresponding layers in the encoder and decoder paths. At each stage of the decoder, the feature maps from the encoder at the same spatial resolution are concatenated with the upsampled decoder features. This allows the network to preserve high-resolution spatial information that might otherwise be lost during downsampling, improving the reconstruction of fine-scale structures. Unlike the residual connections in SRDRN, which use element-wise summation, the UNet skip connections use concatenation, providing the decoder with access to the full set of encoder features.

To efficiently incorporate temporal information,  this architecture has been extended by adding a temporal module similarly as in SRDRN. The temporal features, obtained from the temporal module, are concatenated with the feature maps at the beginning of the decoder, where the spatial resolution is low. This placement allows the temporal information to be propagated through the upsampling path, enabling the network to modulate both coarse and fine-scale spatial features and effectively learn spatio-temporal interactions across the entire high-resolution output.

The complete time-aware UNet is constructed as follows. An initial bilinear upsampling layer is first applied to bring the coarse-resolution input to the desired target resolution. This ensures that the encoder and decoder paths operate at the same spatial scale, allowing skip connections to effectively transfer information between the corresponding encoder and decoder layers. After this, $n_K$ initial feature extraction convolutional layers with ReLU activation are used to obtain rich, higher dimensional, low-level representation. The features are then passed through a sequence of $n_M$ encoder blocks, each consisting of two convolutional layers with ReLU activation and a max pooling operation. Each encoder block reduces the spatial resolution by a factor of two while increasing the number of feature channels. The outputs of the encoder blocks are stored for the skip connections.

\begin{figure}[h]
    \centering
    \includegraphics[width=0.8\linewidth]{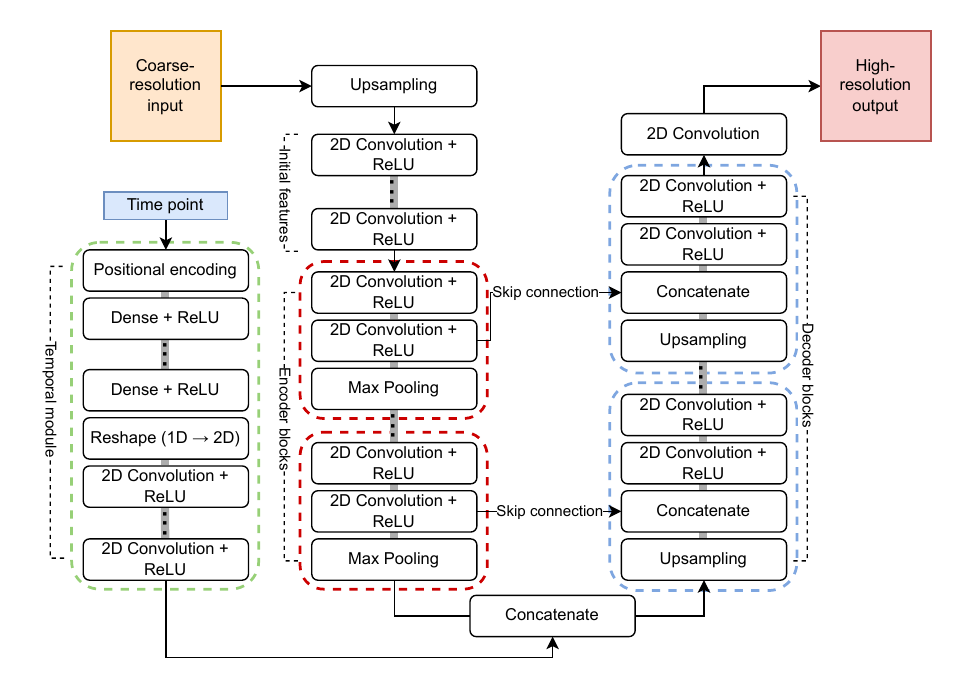}
    \caption{\label{figSchemeUnet}A schematic representation of time-aware UNet. The temporal module is presented within dashed green line, the encoder blocks within dashed red line and decoder blocks within dashed blue line.}
\end{figure}

At the bottleneck, two convolutional layers with ReLU activation are applied to obtain the final encoded spatial features, after which an optional dropout layer is applied and the features obtained from the temporal module are concatenated. The concatenated spatial and temporal features are fed into the decoder. The decoder is symmetric to the encoder and therefore consists of $n_M$ decoding blocks, each containing a bilinear upsampling layer followed by two convolutional layers with ReLU activation. The skip connections are applied by concatenating the features from the encoder blocks with the corresponding decoder features at each spatial scale. Finally, a convolutional output layer is applied to generate the high-resolution output field. A schematic overview of the time-aware UNet is provided in Fig.~\ref{figSchemeUnet}.

The parameters of the UNet network are trained by minimizing the MSE between the estimated high-resolution output and the true high-resolution field. The main hyperparameters include the number of encoder and decoder blocks $n_M$, the number of initial feature extraction layers $n_K$, the number of channels within each convolution layer, and temporal encoding process and the number of hidden layers within it. In addition, general hyperparameters such as the optimizer, learning rate, batch size and the number of epochs must be specified.

In practice, UNet architectures for spatial downscaling typically employ two to four encoder–decoder blocks ($n_M = 2$–$4$), each halving and doubling the spatial resolution in the encoder and decoder paths, respectively. The number of initial feature extraction layers $n_K$ is usually one or two, allowing the network to learn low-level spatial representations before entering the encoder path. The number of channels is often doubled at each deeper encoder level and symmetrically reduced in the decoder, which allows the skip connections between encoder and decoder paths. The temporal module and general hyperparameters are selected as in the time-aware SRDRN.

\section{R package %\texttt{SpatialDownscaling} 
for spatial downscaling}
\label{sec:rpackage}

As part of this work, all considered spatial downscaling methods have been implemented in R package \texttt{SpatialDownscaling}\footnote{\url{https://github.com/mikasip/SpatialDownscaling}}. The main functions of the package are \texttt{srdrn} and \texttt{unet} which implement the SRDRN and UNet architectures as described in Sections \ref{sec:srdrn} and \ref{sec:unet}, respectively. Both functions support optional temporal conditioning through a shared temporal module architecture. The methods are built mainly on top of the packages \texttt{keras3} \citep{keras3} and \texttt{tensorflow} \citep{tensorflow}.

Both \texttt{srdrn} and \texttt{unet} follow a consistent interface, accepting coarse-resolution input data and fine-resolution target data as multi-dimensional arrays with dimensions $[\text{x}, \text{y}, \text{time}]$. The spatial dimensions x and y represent the grid coordinates, while the time dimension corresponds to different temporal snapshots. Optional validation data can be provided with the parameters \texttt{val\_coarse\_data} and \texttt{val\_fine\_data}. The most important parameters controlling the network architectures correspond directly to the mathematical formulations presented in the Sections \ref{sec:srdrn} and \ref{sec:unet}.

Temporal conditioning is activated by providing the \texttt{time\_points} parameter, which is a vector of time indices corresponding to each temporal snapshot in the data. When temporal information is included, the temporal module parameters control how temporal patterns are encoded. The \texttt{cyclical\_period} parameter specifies the period for cyclical temporal patterns (e.g., 365 for daily data with annual cycles), while \texttt{temporal\_basis} controls the number and placement of radial basis functions for temporal encoding (default: \texttt{c(9, 17, 37)}). The \texttt{temporal\_layers} parameter defines the hidden layer sizes in the initial temporal feed-forward network (default: \texttt{c(32, 64, 128)}) and the parameters \texttt{temporal\_cnn\_filters} (default: \texttt{c(8, 16)}) and \texttt{temporal\_cnn\_kernel\_sizes} (default: \texttt{list(c(3, 3), c(3, 3))}) define the number of temporal CNN layers and their filters and kernel sizes, respectively. Alternatively, users can set \texttt{cos\_sin\_transform} to \texttt{TRUE} to use sinusoidal transformation instead of radial basis functions for cyclical time representation. If the validation data are provided, the user should also provide the validation time points (\texttt{val\_time\_points}) if the temporal modeling is included.

For the SRDRN architecture, the key structural parameter is \texttt{num\_residual\_blocks}, which controls the number of residual blocks $n_R$ in the network (default: 3). The number of upsampling blocks is automatically calculated as $\lfloor \log_2(r) \rfloor$, where $r$ is the spatial upscaling factor determined from the ratio of fine to coarse resolution dimensions. The parameter \texttt{num\_res\_block\_filters} defines the number of filters in the convolutional layers of the residual blocks (default: 64), while the vector \texttt{upscaling\_filters} defines the number of filters used in each upscaling block, with defaults taken sequentially from \texttt{c(64, 32, 16, 8, 4, 2)} based on the number of upscaling blocks needed.

For the UNet architecture, \texttt{initial\_filters} specifies the number of filters in the $n_K$ initial feature extraction layers (default is 16), while \texttt{filters} defines the number of filters in each of the $n_M$ encoder/decoder blocks (default: \texttt{c(32, 64, 128)}). The \texttt{kernel\_sizes} parameter controls the convolutional kernel sizes for each encoder/decoder block.

Both functions share common training parameters including \texttt{learning\_rate} for the Adam optimizer (default: 0.001), \texttt{epochs} for the number of training iterations (default is 10), and \texttt{batch\_size} for mini-batch size of the stochastic gradient descent (default: 32). Loss function, optimizer and nonlinear activation functions can be modified with parameters \texttt{loss} (default: ``mse''), \texttt{optimizer} (default: ``adam'') and \texttt{activation} (default: ``relu''), respectively. The options for the parameters are inherited from the \texttt{keras3} package.

The typical workflow for spatial downscaling by using the package, consists of three main steps:
\begin{enumerate}
\item Prepare coarse and fine resolution training data as arrays with dimensions $[\text{x}, \text{y}, \text{time}]$
\item Train a model by calling \texttt{srdrn} or \texttt{unet} with the training data and desired parameters
\item Generate high-resolution predictions for new coarse-resolution inputs using the  \texttt{predict} method
\end{enumerate}

For example, to train a time-aware SRDRN model with coarse training data of dimensions $32 \times 32 \times 1000$ and fine training data of dimensions $128 \times 128 \times 1000$ (corresponding to a 4$\times$ upscaling factor), one would call \texttt{srdrn} with the input and target data arrays, provide the time points vector (e.g., 1 to 1000), specify a cyclical period of 365 for annual cycles, and set the desired number of residual blocks and training epochs. The resulting model object can then be used with the \texttt{predict} function to generate downscaled predictions for new coarse-resolution data by providing both the new coarse data array and corresponding time points. The same workflow applies to UNet training using the \texttt{unet} function, with appropriate architectural parameters such as the filter sizes for encoder/decoder blocks.

Both methods handle missing values automatically by masking them during training and prediction. The methods use \texttt{keras3} and \texttt{tensorflow} backend for efficient computation. Model objects returned by the training functions contain the trained neural network, normalization parameters, and metadata required for prediction, allowing trained models to be saved and reused later for inference.

In addition to SRDRN and UNet methods, the package contains an implementation for a statistical baseline model BCSD. The method is called similarly as UNet and SRDRN methods by giving coarse and fine training datasets. The key parameters include \texttt{method} for specifying the interpolation approach (default: bilinear), \texttt{n\_quantiles} for controlling the number of quantile levels used in the quantile mapping (default: 100), and \texttt{reference\_period} for restricting the quantile calculation to a specific temporal subset of the training data. The \texttt{extrapolate} parameter (default: TRUE) determines whether values outside the training range are extrapolated using constant shift corrections, while \texttt{normalize} controls data standardization (default: TRUE). Unlike the neural network approaches, BCSD does not require iterative training and instead creates a quantile mapping function directly from the training data, making it computationally efficient but usually less flexible in capturing complex spatial patterns. The \texttt{predict} method for BCSD objects applies the learned quantile mapping followed by spatial interpolation to generate high-resolution fields.

\section{Case study}\label{casestudy}

Tropospheric ozone is a key air pollutant with strong spatio-temporal variability, influencing both human health and ecosystem functioning. High-resolution ozone fields are essential for accurately assessing local air quality, but the satellite products are often provided only at coarse resolution level. In this study, the focus is on spatial downscaling of daily gridded satellite-derived ozone concentration data over Italy. Two well-known image-to-image architectures, SRDRN and UNet, are compared with their proposed time-aware variants. For the time-aware variants, both sinusoidal and RBF encodings are considered. In addition, the BCSD method is included as a statistical baseline for comparison.

The analysis is performed in R 4.4.1 \citep{Rcoreteam} and all implemented methods are available in the R package \texttt{SpatialDownscaling} described in Section \ref{sec:rpackage}. 
The reproducible code to replicate all results in the case study is provided in GitHub\footnote{\url{https://github.com/mikasip/OzoneSpatialDownscaling}}.

\subsection{Analyzed data} 

This study focuses on the spatial downscaling of gridded tropospheric  $\text{O}_3$ concentrations ($\mu g/m^3$) at ground level, over the entire Italian territory. 
High-resolution ozone fields are essential for accurately assessing local air quality, particularly in regions characterized by challenging landscapes and intricate meteorological dynamics. The target area includes the whole national territory, encompassing diverse geographical features (from high Alpine and Apennine systems to coastlines and the extensive plains). Italy represents a key economic and demographic hub in Southern Europe, whose importance is intrinsically linked to its environmental complexity.
The most persistent pollution issues of the territory are driven by several factors, notably the distinct geographic basin effect experienced in large flat areas such as the Po Valley (Pianura Padana), as well as in other densely populated or industrialized areas distributed across the peninsula and coastal zones. This configuration, bordered by the Alps to the north and the Apennines to the south, critically traps pollutants and restricts atmospheric mixing. Consequently, in those areas, $\text{O}_3$ concentrations often exceed established health and environmental thresholds, especially during the summer. 
In particular,  elevated contamination in several Italian regions is due to a confluence of conditions: high levels of precursor pollutants (such as nitrogen oxides from vehicle traffic and industrial operations), intense solar radiation and region-specific weather patterns.

\begin{figure}[htbp]
    \centering
    \begin{subfigure}[t]{0.32\textwidth}
        \centering
        \includegraphics[width=\linewidth]{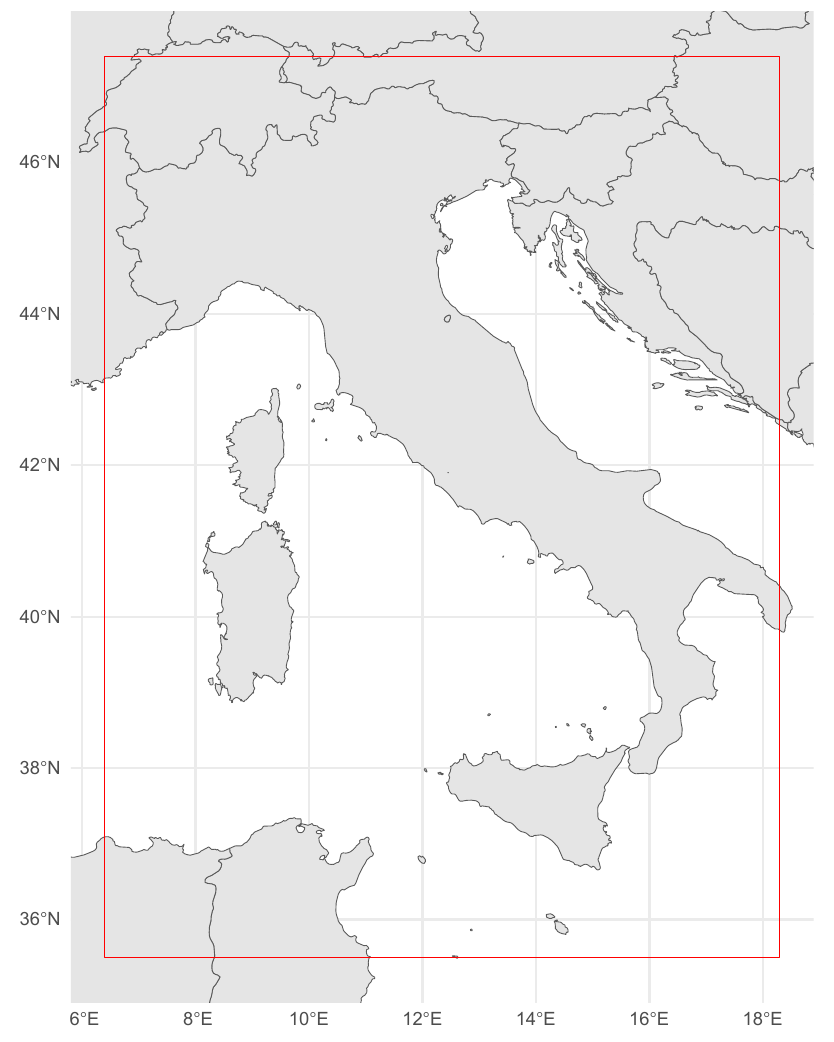}
          \begin{flushleft}
           \hspace{.3cm}    a)
          \end{flushleft}
    \end{subfigure}%
    \begin{subfigure}[t]{0.34\textwidth}
       \centering
        \includegraphics[width=\linewidth]{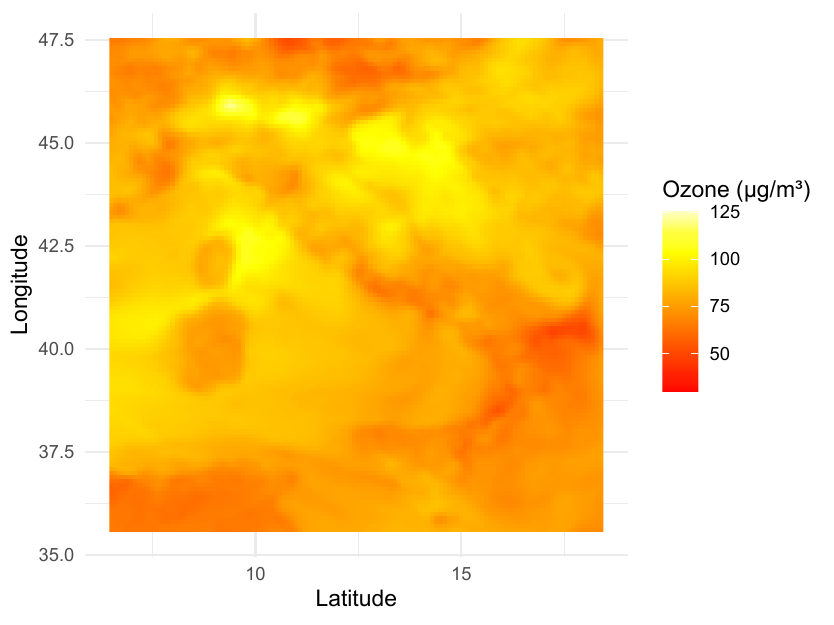}
              \begin{flushleft}
              \hspace{.65cm}   b)
          \end{flushleft}
    \end{subfigure}%
    \hfill
    \begin{subfigure}[t]{0.34\textwidth}
        \centering
        \includegraphics[width=\linewidth]{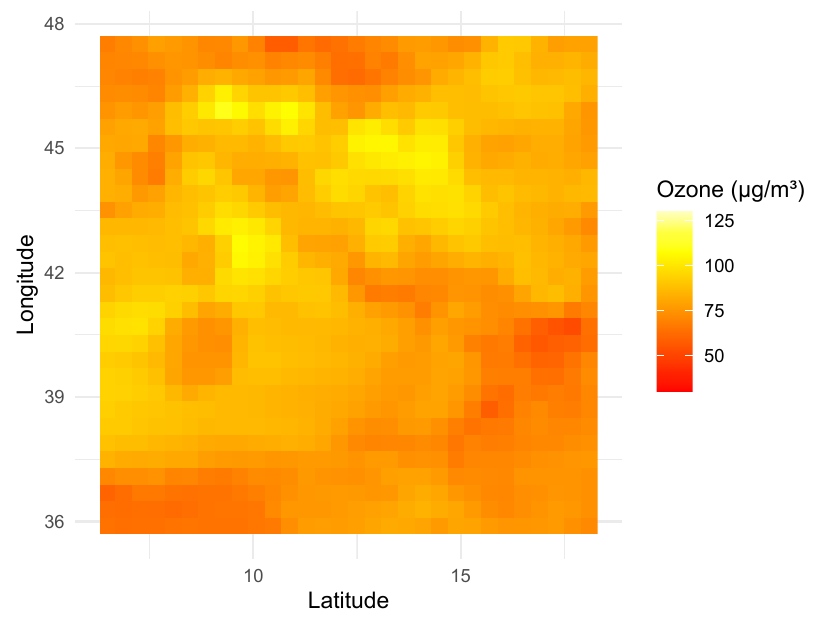}
          \begin{flushleft}
             \hspace{.65cm}    c)
          \end{flushleft}
    \end{subfigure}
    \caption{\label{fig:study_area}The map of the study area is marked with red square (a). The satellite-derived ozone levels are provided at 1/7/2022 at fine-resolution (b) and at coarse-resolution (c).}
\end{figure}

A visual overview of the study area and an example of ozone distributions at different spatial resolutions are provided in Figure \ref{fig:study_area}a–c.

The gridded $\text{O}_3$ concentrations at ground level were sourced from the {Copernicus Atmosphere Monitoring Service (AQ-CAMS)}. The daily data span six years, from 2018 to 2023, yielding a total of 2,191 records. The input dataset consisted of low resolution $\text{O}_3$ data ($0.4^{\circ}$ on a $30\times 30$ grid), obtained by averaging the original high resolution fields ($0.1^{\circ}$ on a $120 \times 120$ grid) over non-overlapping $4 \times 4$ spatial blocks.\\
The objective of the study is to reconstruct the high-resolution ozone fields from their coarse counterparts over time.

The performance of the proposed approach is evaluated using this satellite gridded tropospheric ozone dataset. The compared methods include variants of SRDRN and UNet, both with and without positional encoding.
For model evaluation, the last 200 time points are reserved as test data. 

Figure \ref{fig:study_area}a–c show the study area and examples of satellite-derived ozone concentration maps at fine and coarse resolutions for 1 July 2022.
The figure highlights the strong contrast between fine and coarse-scale representations: fine-resolution data capture local gradients and hotspots (particularly in urbanized or industrial regions and over challenging landscapes), while coarse-resolution fields smooth out these spatial variations. This difference visually demonstrates the importance of this downscaling approach for recovering small-scale features crucial to accurate air quality assessment across Italy.

\subsection{Model parameters} 
In all SRDRN variants, $n_L = 3$ encoder blocks and $n_B = 2$ upsampling blocks with upscale factors $r_1, r_2 =2$ are used. The convolutional layers of the encoder blocks have 64 filters, while the upsampling blocks have 64 and 32 filters, respectively. In the temporal modules of the time-aware UNet and SRDRN models, the initial feed-forward network consists of layers with 32, 64, 128, 256 and 900 hidden units, respectively. The last hidden layer has 900 hidden units in order to reshape the output into $30 \times 30$ array, which is the size of the coarse-resolution input. Then, two consecutive convolutional layers with 8 and 16 filters, respectively, are applied to produce a representative spatial representation of the temporal information.

For RBF variants, the radial basis functions are constructed based on seasonal time period $t=1,\dots, 365$ instead of the absolute time point to allow the model to generalize better to future values. The resolution levels $G=(9, 17, 37)$ are used to create the seasonal RBFs.

In each UNet variant, one initial convolutional layer with 16 filters and two encoding blocks with filter sizes 32 and 64 are used. The bottleneck convolutional layers have 128 filters each and the decoder is symmetric to the encoder, having 64 and 32 filters in the two decoding blocks.

\begin{figure}[htbp]
    \centering
    \begin{subfigure}[t]{0.5\textwidth}
        \centering
        \includegraphics[width=\linewidth]{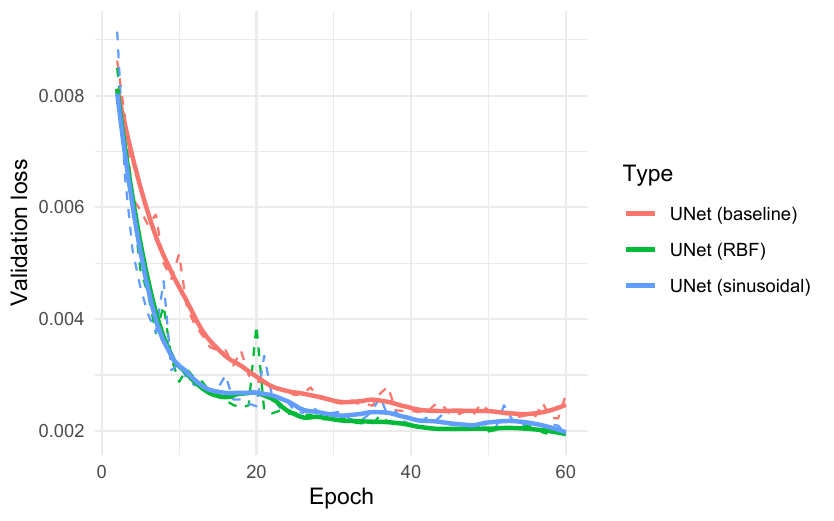}
    \end{subfigure}%
    \begin{subfigure}[t]{0.5\textwidth}
       \centering
       \includegraphics[width=\linewidth]{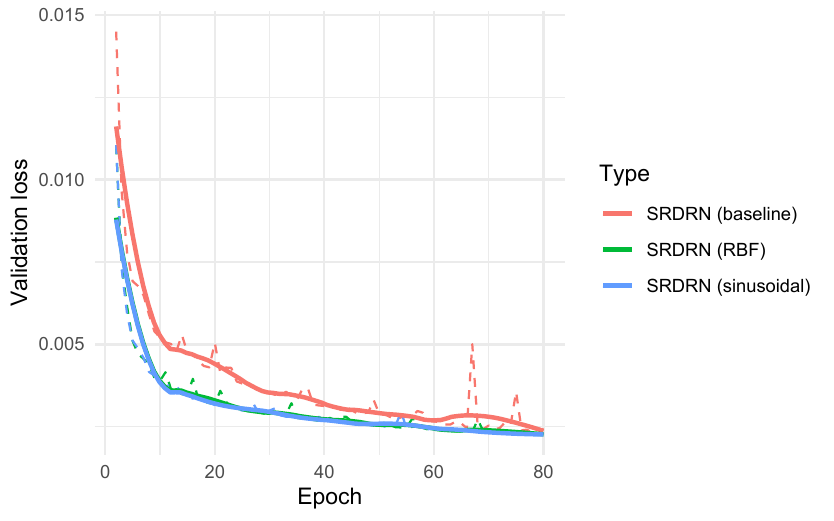}
    \end{subfigure}%
    \caption{The comparison of the validation losses against the number of training epochs for the baseline UNet (a) and SRDRN (b) models along with their time-aware extensions. The dashed lines represent the exact validation losses, while the solid lines are smoothed using locally weighted regression (LOESS) with smoothing span 0.3.}
    \label{fig:convergences}
\end{figure}

The number of training epochs for SRDRN and UNet was determined based on an initial training where a validation accuracy was monitored. In validation step, the models were trained on 1500 first time points of the training dataset, and evaluated based on the remaining 490 time points. The convergence and good generalization to unseen data was ensured by selecting the number of epochs at which the validation accuracy stabilizes. The convergence curves are provided in Fig. \ref{fig:convergences}. As the validation accuracy for SRDRN methods are steadily decreasing and stabilize only right before the final training epoch, the final models are trained for the total of 80 epochs. Because the UNet models converge more rapidly, and stabilize at around epoch 40, the final models are trained for the total of 40 epochs. 
Moreover, based on the validation losses per training epoch, the time-aware variants of UNet and SRDRN achieve lower validation loss and converge significantly faster than their baseline models.
The learning rate was set to 0.0005 for all models in validation step and in final training.
%\sandra{NOTE: I would add some additional comments on the training and validation steps}

\subsection{Performance metrics} 
The performances of the different downscaling methods are evaluated using mean absolute error (MAE), root mean squared error (RMSE), Kling-Gupta efficiency (KGE) developed by \cite{gupta2009decomposition}. MAE and RMSE are standard error measures that quantify the average magnitude of prediction errors, with lower values indicating better performance. KGE provides a measure that decomposes the model performance into three components: correlation, bias and variability. It is defined as
\begin{align}
    \text{KGE} = 1 - \sqrt{(r-1)^2 + (\alpha - 1)^2 + (\beta - 1)^2},
\end{align}
where $r$ is the linear correlation coefficient between the predictions and the true values, $\alpha = \sigma_{\hat{x}} / \sigma_x$ is the variability ratio with $\sigma_{\hat{x}}$ and $\sigma_x$ denoting the standard deviations of predicted and observed values, respectively, and
$\beta = \mu_{\hat{x}} / \mu_x$ is the bias ratio with $\mu_{\hat{x}}$ and $\mu_x$ denoting their means. KGE ranges from negative infinity to 1, where the optimal value of 1 indicates a perfect fit for the model. The value greater than $-0.41$ indicates performance of mean flow \citep{knoben2019inherent}, meaning that the true variables would be estimated by their mean.

\subsection{Results} 
After initial validation steps, the different models are trained by using the whole training dataset. The performance metrics, presented in Table~\ref{tab:performances}, are calculated for the test data set.

\begin{table}[htbp]
\caption{\label{tab:performances}Performance metrics (MAE, RMSE and KGE) for different methods in spatial downscaling of ozone in Italy.}
    \centering
    {\small\begin{tabular}{lllll}
          \hline
          Method & MAE & RMSE & KGE \\
         \hline
         BCSD & 1.453 & 2.165 & 0.9875 \\
         SRDRN (baseline) & 0.765 & 1.076 & 0.9924 \\
         SRDRN (sinusoidal) & 0.677 & 0.965 & 0.9982 \\
         SRDRN (RBF) & 0.691 & 0.973 & 0.9932 \\
         UNet (baseline) & 0.782 & 1.080 & 0.9809 \\
         UNet (sinusoidal) & 0.659 & 0.950 & 0.9925 \\
         UNet (RBF) & 0.640 & 0.929 & 0.9970 \\\hline
    \end{tabular}}
\end{table}

All SRDRN and UNet variants substantially outperformed BCSD. Among the baseline methods without temporal module, SRDRN reached lower errors and higher KGE than UNet. Inclusion of the temporal information consistently improved the performance for both architectures. For SRDRN, the sinusoidal positional encoding yielded the best results, whereas in UNet, RBF encoding was slightly better. The relative MAE improvements of the time-aware methods compared to the baseline methods were 11.5\% for sinusoidal SRDRN, 9.7\% for RBF SRDRN, 15.7\% for sinusoidal UNet and 18.2\% for RBF UNet. The differences between sinusoidal and RBF encoding were small, yet both variants outperformed their baseline counterparts and provided faster convergence compared to the baseline methods. Overall, the time-aware UNet with RBF encoding yielded the highest performance across all metrics.
%\sabrina{The absolute prediction errors for the SRDRN variants at time points 1, 100, and 200 of the test set are detailed in Fig. \ref{fig:placeholder1} , while the same analysis for the UNet variants is presented in Fig. \ref{fig:placeholder2}.}
%\begin{figure}[htbp]
%    \centering
%    \includegraphics[width=1\linewidth]{srdrn_errors.pdf}
%    \caption{\label{fig:placeholder1}The absolute prediction errors for the SRDRN variants at time points 1 (first row), 100 (second row) and 200 (third row) of the test set. The first column presents the errors for the baseline SRDRN, the second for time-aware SRDRN with RBF encoding and the third for time-aware SRDRN with sinusoidal encoding.\sabrina{\textbf{NOTE: Could you please improve the resolution of Figure?}}}
%\end{figure}
%\begin{figure}[htbp]
%    \centering
%    \includegraphics[width=1\linewidth]{unet_errors.pdf}
%    \caption{\label{fig:placeholder2}The absolute prediction errors for the UNet variants at time points 1 (first row), 100 (second row) and 200 (third row) of the test set. The first column presents the errors for the baseline UNet, the second for time-aware UNet with RBF encoding and the third for time-aware UNet with sinusoidal encoding. \sabrina{\textbf{NOTE: Could you please improve the resolution of Figure?}}}
%\end{figure}

From a computational perspective, the temporal module adds negligible overhead. It consists of a simple feed-forward network together with an optional stack of light weight convolutional layers, while the computationally demanding operations remain in the main SRDRN and UNet modules. Consequently, the training times per epoch for baseline and time-aware variants were nearly identical, and no practical differences in runtime were observed. However, because of the faster training convergence of the time-aware models, they can be trained with lower computational resources than the baseline models. 

\begin{figure}[htbp]
    \centering
    \includegraphics[width=1\linewidth]{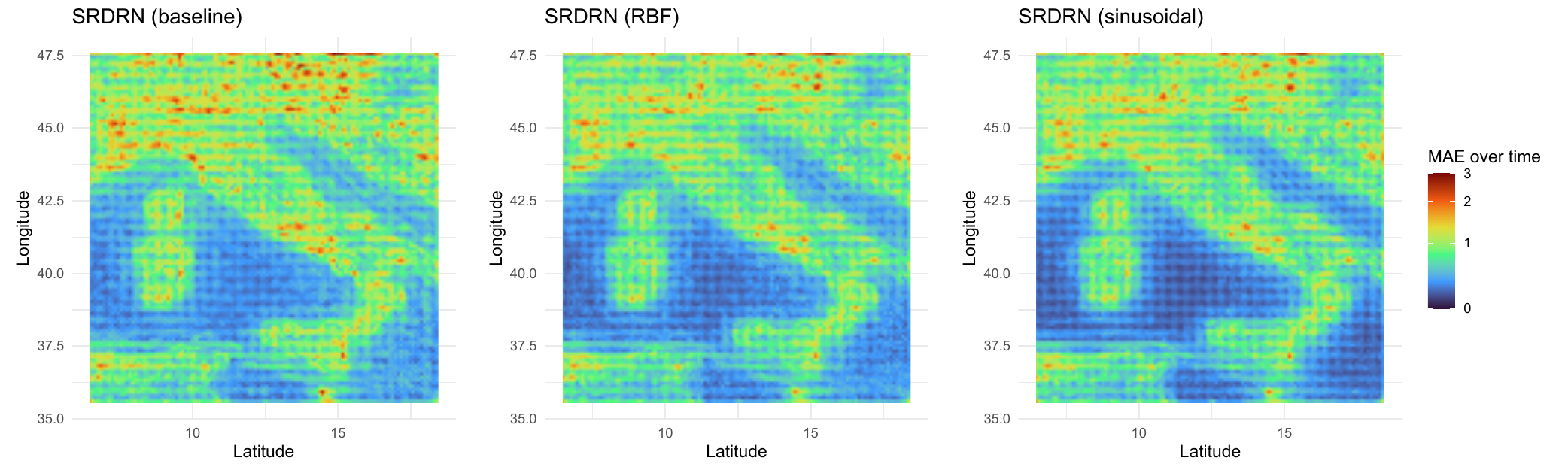}
    \caption{\label{fig:placeholdertime1}The mean absolute prediction errors over the test set for the SRDRN variants. The first column presents the errors for the baseline SRDRN, the second for time-aware SRDRN with RBF encoding and the third for time-aware SRDRN with sinusoidal encoding.}
\end{figure}

\begin{figure}[htbp]
    \centering
    \includegraphics[width=1\linewidth]{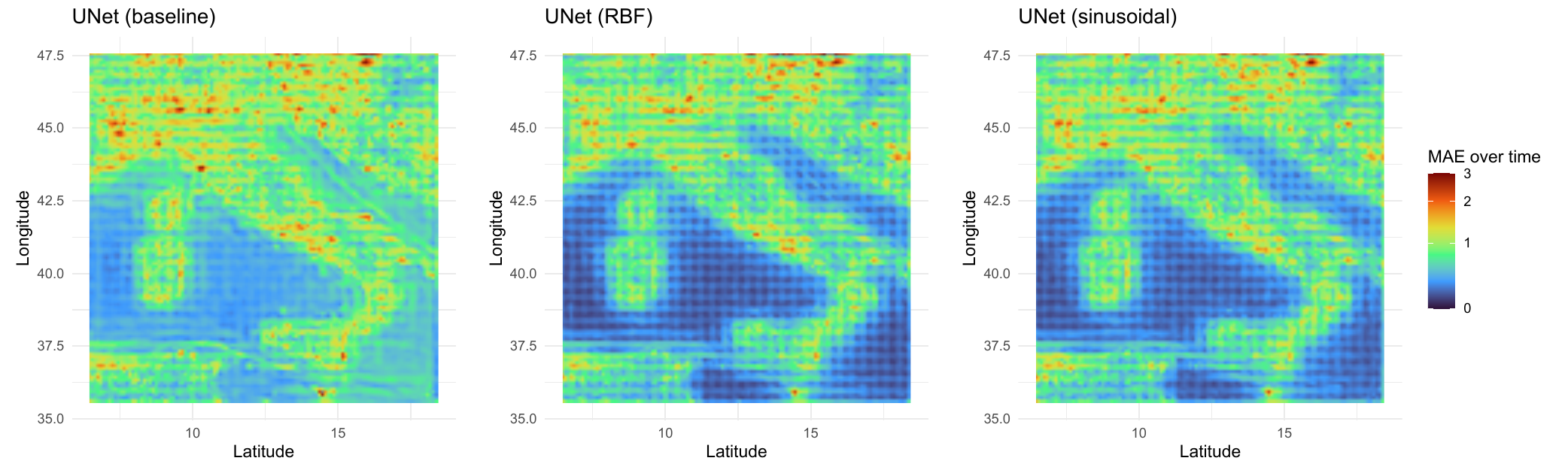}
    \caption{\label{fig:placeholdertime2}The mean absolute prediction errors over the test set for the UNet variants. The first column presents the errors for the baseline UNet, the second for time-aware UNet with RBF encoding and the third for time-aware UNet with sinusoidal encoding.}%\sabrina{NOTE: Could you please improve the resolution of Figure?}}
\end{figure}

The performance of the methods is further compared by calculating average MAE spatial maps over the test time period. The spatial patterns are presented in Fig. \ref{fig:placeholdertime1} for SRDRN and Fig. \ref{fig:placeholdertime2} for UNet. While the maps show only slightly lower errors for time-aware models in Southern and Central Italy, more clear differences are present particularly across the Northern Italy. For all methods, the lowest errors are in the sea whereas the highest errors are present in the Northern Italy.


%\sandra{NOTE: I would add some relative measures of improvement, figures of errors, is there any systematic behaviour? what happens on the the last 200 time points that are reserved as test data? do you predict on a finer grid and for these additional time point?}
\section{Conclusions and discussion}\label{concldis}

In this study, time-aware extensions of SRDRN and UNet architectures were implemented by incorporating a temporal module that extracts information from positionally encoded time points. The resulting models were evaluated for the spatial downscaling of ozone concentrations in Italy. 

This downscaling process is critical because satellite observations of $\text{O}_3$ are typically available only at coarse resolution (e.g. $0.4^{\circ}$ resolution), which is insufficient for local-scale exposure assessments and policy decisions. The specific case study focused on reconstructing high-resolution fields ($0.1^{\circ}$ resolution) across the Italian territory, a vast area characterized by high $\text{O}_3$ levels due to high precursor pollutants, intense solar radiation, and a distinctive geographical configuration.

All deep learning models considered substantially outperformed the statistical baseline BCSD. Across all metrics, the time-aware variants consistently outperformed their baseline counterparts in terms of prediction accuracy, while introducing only marginal computational overhead. The time-aware extensions also provided considerably faster training convergence compared to the baseline methods.

%\sabrina{This efficiency is achieved because the temporal module is structured as a {simple feed-forward network}, ensuring that the training times per epoch for baseline and time-aware variants were nearly identical.} Differences between sinusoidal and RBF positional encodings were small. However, sinusoidal encoding provided the best results for UNet, whereas RBF was slightly more effective in SRDRN. \sabrina{The temporal modules provide a significant boost in accuracy:
%a) for the UNet architecture, the time-aware UNet (sinusoidal) achieved the best overall performance, reducing the MAE by 5.8\% (from 0.779 to 0.734) and the RMSE by 5.2\% (from 1.174 to 1.113), while attaining the highest KGE (0.9980); b) for the SRDRN architecture, the RBF encoding yielded the greatest improvement, reducing the MAE by 6.3\% (from 0.806 to 0.755) and the RMSE by 5.4\% (from 1.221 to 1.155), with a strong KGE of 0.9971.
%}

The spatial patterns of the prediction errors were further studied by spatial maps of MAEs averaged over the test time period for the different methods. Although the differences are modest, the average MAEs of the time-aware variants were clearly lower especially across the Northern Italy.

In conclusion, the performance improvements of the time-aware variants highlight the effectiveness and potential of including a lightweight temporal module with positional encoding to incorporate temporal information into convolutional architectures for spatial downscaling. Given the faster convergence and similar training time per epoch, the proposed models can be trained successfully with lower computation time than the baseline models without temporal module.

While both encoding strategies performed well in this study, future work should explore their performance across different datasets and under varying levels of downscaling complexity using simulation studies. In particular, for more challenging scenarios such as higher spatial scaling factors or more complex environmental phenomena, the flexibility of RBF-based encoding may offer an advantage.

%Although the motivation in this work was to computationally light, yet effective way, to include temporal information to common spatial downscaling deep learning architectures, it will be important to benchmark these time-aware methods against more resource-intensive temporal methods, such as recurrent or attention-based networks, which can explicitly leverage temporal dependencies and past observations. 

Although the motivation of this work was to develop a computationally light yet effective way to incorporate temporal information into common spatial downscaling deep-learning architectures, it will be important to benchmark these time-aware methods against more resource-intensive temporal models, such as recurrent or attention-based networks, which can explicitly leverage temporal dependencies and past observations.

\section*{Acknowledgments}
This research was partially supported by the 
European Union-NextGenerationEU with the Cascade Open Calls published by ALMA MATER STUDIORUM - University of Bologna, inside the Project GRINS funded by PNRR - Mission 4, Component 2, Investment 1.3 {\lq\lq Partnership extended to Universities, Research Centers, Firms and research projects funding\rq\rq}, D.D. 341 of 15/03/2022, {\lq\lq ECoST-DATA, Exploring Spatio-Temporal Environmental Conditions: Harmonized Databases and Analytical Techniques\rq\rq}, CUP: J33C22002910001.\\
The work of KN was supported by the Research Council of Finland (363261). The work of MS was supported by the Vilho, Yrjö and Kalle Väisälä foundation. The work of ST was supported by the Research Council of Finland (356484).

\bibliographystyle{elsarticle-harv}
\bibliography{references}

\end{document}